\PassOptionsToPackage{numbers,sort&compress}{natbib}
\documentclass[a4paper,fleqn]{cas-dc}
\usepackage{ragged2e}
\usepackage{natbib}
\usepackage{algorithmic}
\usepackage{algorithm}
\usepackage{graphicx}
\usepackage{pifont}
\usepackage{framed}
\usepackage{multicol} 

\usepackage{nomencl}

\makenomenclature

\renewcommand*\nompreamble{\begin{multicols}{2}}

\renewcommand*\nompostamble{\end{multicols}}
\def\tsc#1{\csdef{#1}{\textsc{\lowercase{#1}}\xspace}}
\tsc{WGM}
\tsc{QE}
\tsc{EP}
\tsc{PMS}
\tsc{BEC}
\tsc{DE}

\begin{document}
\let\WriteBookmarks\relax
\def\floatpagepagefraction{1}
\def\textpagefraction{.001}

\shorttitle{Robot Nurse in Infectious Pandemic Management}

\shortauthors{Sifat MMH et~al.}

\title [mode = title]{Cost-Optimized Systems Engineering for IoT-Enabled Robot Nurse in Infectious Pandemic Management}                      


%
\author[1]{Md Mhamud Hussen Sifat}[orcid=0009-0000-9579-9596]
\author[1]{Md Maruf}[orcid=0000-0001-8627-5512]
\cormark[1]                         
\ead{maruf.mte.17@gmail.com}       

\cortext[1]{Corresponding author.}
\author[1]{Md Rokunuzzaman}
\affiliation[1]{organization={Rajshahi University of Engineering \& Technology},
    city={Rajshahi},
     citysep={}, 
    postcode={6204}, 
    country={Bangladesh}}

\begin{abstract}
The utilization of robotic technology has gained traction in healthcare facilities due to the progress made in this field which enables time and cost savings, minimizes waste, and improves patient care. Digital healthcare technologies that leverage automation such as robotics and artificial intelligence have the potential to enhance the sustainability and profitability of healthcare systems in the long run. However, the recent COVID-19 pandemic has amplified the need for cyber-physical robots to automate check-ups and medication administration. A Robot nurse is a robot that is controlled by the Internet of Things (IoT) and can serve as an automated medical assistant while also allowing for supervisory control based on custom commands. This system aids in eliminating the risks of infection and provides effective results in handling pandemics. This research puts forth a test case for the management of such situations with a nurse robot that possesses advanced intelligence to check a patient's health status and take action accordingly. Furthermore, the study assesses the efficacy of the system in terms of medication administration, health status, and system life-cycle.
\end{abstract}


\maketitle
\section{Introduction}
The medical industry is witnessing a steady rise in emergencies, with a diverse array of diseases affecting individuals. However, there is a notable shortage of medical personnel available to provide assistance. This scarcity becomes even more acute in the face of a pandemic, as exemplified by the unprecedented impact of Covid-19. The world has witnessed firsthand the immense challenges posed by the sudden emergence of unknown diseases \cite{chari2021impact}.

The impact of COVID-19 has been felt across all continents, with varying degrees of severity. In Asia, which saw the initial outbreak, millions have been affected, and a substantial number lost their lives due to the virus. Europe experienced significant outbreaks in various countries, leading to a high number of infections and fatalities. North America, particularly the United States, faced waves of infections, resulting in a significant toll on both cases and deaths. In South America, the virus spread rapidly, with countries like Brazil and Peru being particularly hard hit. Africa, while initially showing lower infection rates, faced challenges due to healthcare infrastructure \cite{genereux2020one}. The toll on healthcare personnel has been profound, with many frontline workers facing higher risks of infection. Tragically, a significant number of healthcare workers have lost their lives in the line of duty, in the battle against COVID-19 \cite{ulupinar2024intention}. The impact of the pandemic is more critical in third-world countries due to the dense unaware population. Also due to the poor healthcare administration of these countries, the healthcare personnel suffer a lot. These issues can be solved with today's technology where robots can be used to take over such emergencies reducing the chance of affecting healthcare personnel \cite{bandyopadhyay2020infection}.  

In the backdrop of the COVID-19 pandemic, which emerged in a period of notable advancements in global healthcare, it became evident that unforeseen challenges persist. Anticipating and preparing for future health crises requires innovative and sophisticated solutions \cite{murphy2020applications}. With this objective in mind, the development of a robot nurse has been undertaken. This autonomous system is equipped with the capability to conduct comprehensive health assessments, dispense medication, and administer appropriate fluids to patients. Beyond routine checkups, the robot nurse is designed to operate under the guidance of supervisory commands facilitated through the Internet of Things (IoT) framework.

In the development of the robot nurse tailored for pandemic conditions, a systems engineering approach is being employed to ensure a comprehensive and efficient design process \cite{zaitceva2022methods}. This approach is chosen for its systematic and structured methodology, which allows for the integration of various components and functionalities with precision \cite{haberfellner2019systems}. Given the critical nature of the pandemic response, the design constraints of cost and accuracy hold paramount importance. The system engineering approach facilitates a thorough analysis of trade-offs among potential configurations, striking a balance between the need for affordability and the critical demand for high precision in healthcare delivery \cite{lahijanian2018resource}. This method ensures that every element of the robot nurse, from its mechanical components to its software algorithms, is meticulously planned and integrated, taking into consideration the interdependencies and performance requirements. Through this approach, we aim to develop a robot nurse that not only meets the stringent cost constraints but also provides the highest level of accuracy in patient care, thus contributing significantly to the healthcare system's response to pandemics \cite{kyrarini2021survey}.  
\newline
The Contributions of this work are as follows:
\begin{itemize}
    \item Integration of IoT and robotics to remotely monitor patients' vital signs, symptoms, and behavior, enabling healthcare providers to closely track their health status without physical contact. 
    \item The system autonomously stores and dispenses medications, ensuring accurate dosages and minimizing the risk of disease transmission through direct human contact.
    \item Making the robot to be cost-effective with a systems engineering approach.
\end{itemize}

\begin{table*}[!t]   

\begin{framed}

\nomenclature{$B0i$}{Brunch nodes for patient-specific IDs, where i = 1, 2 , 3 ,..., 8.}
\nomenclature{$t_i$}{Specific trajectory for a single brunch node.}
\nomenclature{$f^{tj}(t_{n})$}{Trained trajectory model for brunch nodes, $B0i$.}
\nomenclature{$\xi(\lambda_{i})$}{Health condition defining state function containing states$ \lambda_0, \lambda_1, \lambda_2, ...,\lambda_n$.}
\nomenclature{$f^{tj}(t_i^{-1})$}{Inverse trajectory for returning to base station.}
\nomenclature{$\psi(\gamma_i)$}{Trained output function containing outputs $\gamma_0, \gamma_1,$ $\gamma_2,...,\gamma_n$ for different combination of health state function, $\xi(\lambda_{i})$.}
\nomenclature{$\sigma(\rho_i)$}{Actuation output function defined inside the control system containing $\rho_0 ,\rho_1, \rho_2,...,\rho_n$.}
\nomenclature{$S^t_0$}{Real-time command input at t=0.}
\nomenclature{$S^{t}_i$}{Real-time command input for different i, where i=1,2,3,...,n activates specific actuation profiles.}
\printnomenclature

\end{framed}

\end{table*}

\section{Background}
\justifying
Significant advancements have been achieved in the field of nursing robots \cite{kim2023patient}. The surgical robots and teleoperations have become widespread, showcasing substantial progress \cite{farooq2023mri, xu2024feedback, le2018design}. Additionally, there have been notable developments in the creation of testing kits for pandemic management robots. These ongoing advancements serve as a basis for a relative comparison of progress in this domain.

Ruohan Wang et al. conducted research and experimentation on the advancement and operational efficacy of a medical assistive robot (MAR) named CareDo. The primary emphasis of this study was on the telepresence and teleoperation capabilities of the robot, specifically in the context of remote healthcare provision within isolation wards amidst the COVID-19 epidemic. The combination of a web real-time communications solution and a convolutional neural network for expression identification was also deliberated over in relation to the CareDo robot. Additionally, the utilization of a CNN-based facial expression recognition system for the purpose of monitoring emotional fluctuations in patients was explored \cite{wang2023medical}. Łukasik S et al. investigate the perspectives of prospective healthcare practitioners, specifically medical and nursing students, about the utilization of assistive robots in the provision of care for elderly individuals \cite{lukasik2020role}. The findings of the study indicate a predominantly favorable disposition among prospective medical professionals towards assistive robots, a factor that may influence the reception of such technology among older individuals. Cai et al. discussed the utilization of image processing technology in the estimation of the user's armpit position and conducted a comparative analysis with alternative approaches for attitude prediction. The study is undertaken within the framework of global aging in the 21st century, with a specific emphasis on addressing mobility issues faced by older adults and those with disabilities \cite{cai2021artificial}. In their study, Zhengqi Peng provides a comprehensive analysis and discourse on soft rehabilitation and nursing-care robots. The authors delve into several aspects such as the mechanical structures employed, modeling techniques utilized, and control strategies implemented in these robots. The robots are categorized into two distinct groups according to their actuation technology: tendon-driven actuation and soft intelligent material actuation. They examined the prospective avenues and endeavors within the subject, with the objective of providing guidance for the advancement of sophisticated soft rehabilitation and nursing-care robots \cite{peng2019soft}. The author highlights recent significant contributions in the field of robotic manipulation systems, with a specific emphasis on approaches that utilize vision-based techniques. Furthermore, an examination has been conducted on the present condition, the concerns that researchers have tackled during their experiments, the methodologies they have employed, and the appropriate utilization of these models \cite{shahria2022comprehensive}. Ohneberg et al. have shown that the utilization of assistive robotic systems in the field of care predominantly occurs within the framework of developmental and evaluative stages \cite{ohneberg2023assistive}. In their study, Pineau et al. introduced a mobile robotic assistant that was designed to provide aid to elderly people who experience cognitive and physical limitations, while also offering support to nurses in their everyday tasks. In investigations done within an assisted living home, the robot effectively showcased its capacity to independently offer reminders and guidance to senior patients \cite{pineau2003towards}. The comparative analysis of the works is given in Table 1.

\begin{table*}
    \caption{Comparison of Key Findings in Different Literature Sources.}
    \centering
    \begin{tabular}{|p{3cm}|p{3cm}|p{10cm}|}
    \hline
         Ref. & Work area & Functionalities \\
         \hline
         \cite{wang2023medical} & COVID-19 Pandemic & 
             Video chatting and consultation, facial expression recognition, medical supply system
    \\ \hline
         \cite{cai2021artificial} & Home care & 
             Human pose estimation, human joint position, and predictions
         \\ \hline 
         \cite{pineau2003towards} & Nursing homes & 
             Nursing Assistant, Elderly people assistant
        \\ \hline
         \cite{zemmar2020rise} & Surgical Assistant & 
             Remote surgery assistant, Food, drug, and blood supply
          \\ \hline
         \cite{demaitre2020uvd} & COVID-19 pandemic & 
             UV disinfection, cleaning personnel 
          \\ \hline
         Current work & Infectious pandemics & 
              Remote checkups, automated medical supplies, remote communication and physical state visualization, real-time supervisory control (IoT)
         \\
         \hline
    \end{tabular}
    \label{tab:my_label}
\end{table*}

\subsection{Problem Statement}
A versatile and reprogrammable robot is to be designed, prioritizing cost-effectiveness while aiming for optimal precision. This robot will utilize IoT technology for system control and state storage. Its operations will be conducted remotely, with the entire process overseen from a doctor's chamber. The primary focus is on achieving the highest level of accuracy while keeping costs at a minimum.

\section{Robot Nurse: System Design}
The nursing robot will be developed using a systems engineering methodology, which aligns well with the project's design and resource limitations \cite{couturier2014tracking}. Additionally, the robot comprises multiple individual components that must function cohesively. Achieving this requires a methodical approach driven by comprehensive functional analysis. This systematic process encompasses activities like analyzing requirements, selecting configurations, assessing constraints, and conducting system trade-off analyses \cite{henderson2021value}.  
\subsection{Functional Analysis}
 \begin{figure*}
  \centering\includegraphics[width=0.8\textwidth]{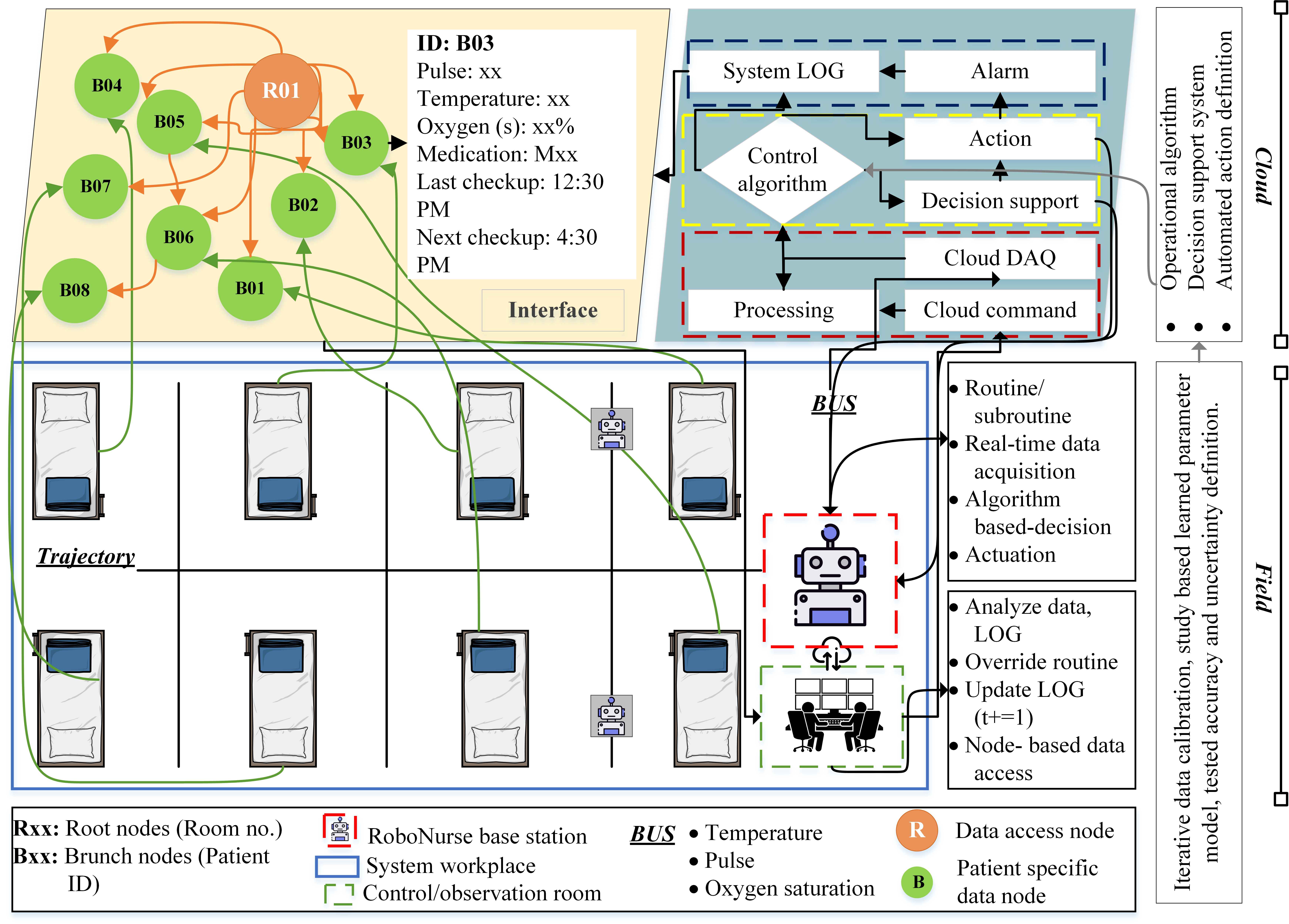}
  \caption{Conceptual framework of the RoboNurse operation and control with the corresponding work environment.}
\end{figure*}%

The complete overview of the robot's operations, routines, and subroutines with its acting environment is shown in Figure 1. The robot works on a specific room accessed remotely by the data end with ID $Rxx$. Further specific patient health status and data log are available with a patient ID $Bxx$. The robot moves from the docking base at a specific checking time or by custom command from the observation room. It follows the trajectory been trained and collects the health state data ($\xi(\lambda_i)$) which is uploaded to the cloud database via the WIFI gateway. The data is processed in real-time to provide the output from the corresponding knowledge base. The medication log is uploaded in parallel with the health state data. Besides, the robot provides supplementary fluids using a robotic arm and places the oxygen mask if needed. The camera module is used to visualize the patient's condition remotely and for frame specification.
\subsection{Constraints}
The robot's design and development have been focused on achieving cost efficiency, which stands as the primary system constraint for the robot nurse. However, another critical constraint arises from the domain of application, namely, the accuracy of data acquisition and medication systems \cite{kyrarini2021survey}. With these considerations in mind, the subsystem analysis entails a trade-off assessment among the modules ($M_i: A-F$) illustrated in Figure 2. By factoring in the primary constraints, a distinct weight matrix can be allocated to gauge the relative significance of the subsystems. However, with these constraints and available options, subsystem components will be chosen from different configurations. The system will further be assessed and if the selected components are available after completing the availability test then the components will be finalized for implementation. 
\begin{equation}
C_f^{c}=\sum_{i=1}^6 (w_c\odot M_i)\hspace{0.1in};Cost\hspace{0.5in}\forall (M_i)
\end{equation}
\begin{equation}
C_f^a=\sum_{i=1}^6 (w_a\odot M_i) \hspace{0.1in};Accuracy\hspace{0.4in}\forall (M_i)
\end{equation}
\begin{equation}
C_f^w=\sum_{i=1}^6 (w_w\odot M_i) \hspace{0.1in};Weight\hspace{0.5in}\in (B,F)
\end{equation}
\begin{equation}
C_f^s=\sum_{i=1}^6 (w_s\odot M_i) \hspace{0.1in};Speed\hspace{0.5in}\in (A,D)
\end{equation}
\begin{equation}
C=\sum_{j=1}^4 C_f^j \longleftarrow Cost-function
\end{equation}
The weight values represent the level of interference the cost function has on a specific module design. These values, displayed in Table 2, are determined by considering the modules' relative significance, adaptability, and their association with a specific weight. These constraint-driven cost functions are considered during the system development process.   

\begin{table}[h]
\centering
\caption{Corresponding weight values (100\%) for the specific subsystem modules ($M_i$)}
\begin{tabular}{|c|c|c|c|c|c|c|}
\hline
 & $M_A$ & $M_B$ & $M_C$ & $M_D$ & $M_E$ & $M_F$\\
\hline
$W_c$ & 10 & 30 & 0 & 10 & 40 & 10\\
\hline
$W_a$ & 20 & 0 & 40 & 35 & 5 & 0\\
\hline
$W_w$ & 0 & 70 & 0 & 0 & O & 30\\
\hline
$W_s$ & 30 & 0 & 30 & 25 & 15 & 0\\
\hline
\end{tabular}
\end{table}

\begin{figure*}
  \centering\includegraphics[width=\textwidth]{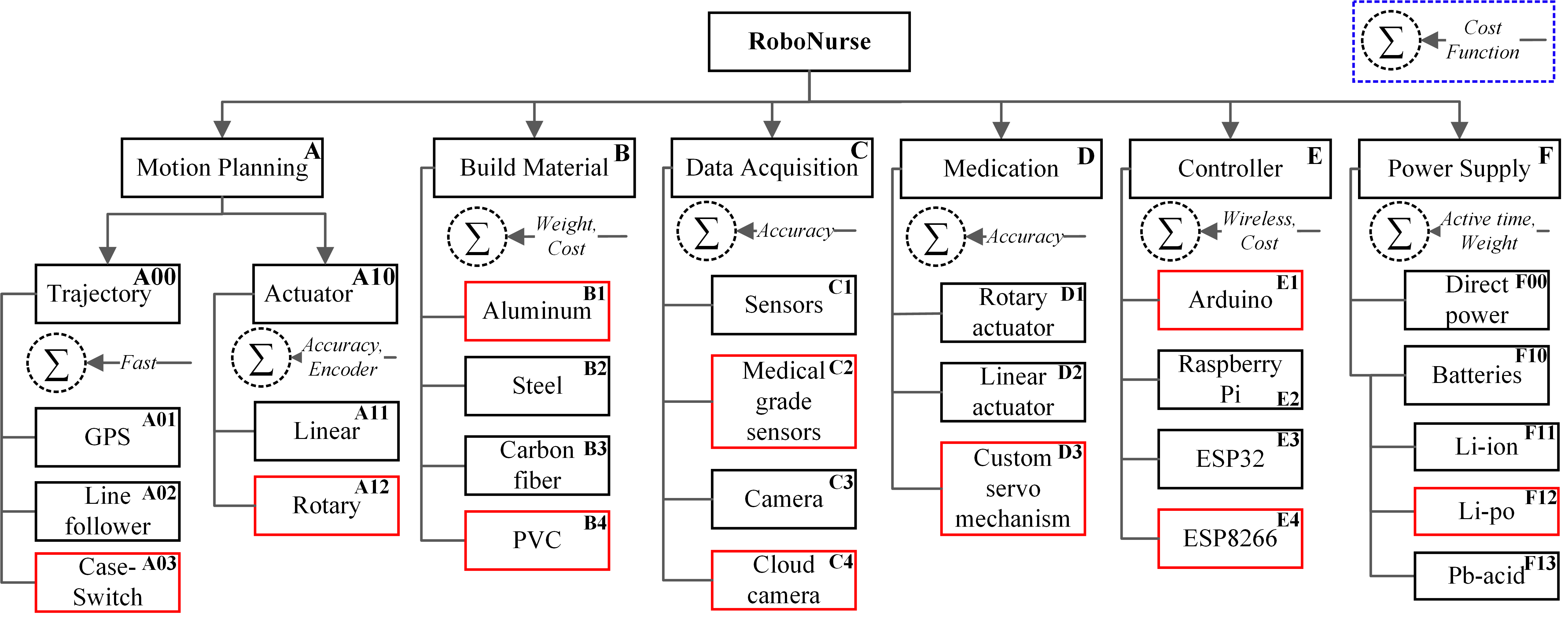}
  \caption{Available alternatives of subsystem configurations with specific design constraints.}
\end{figure*}%

\subsection{System Trade-off}
\begin{figure}
  \centering\includegraphics[width=3.5in]{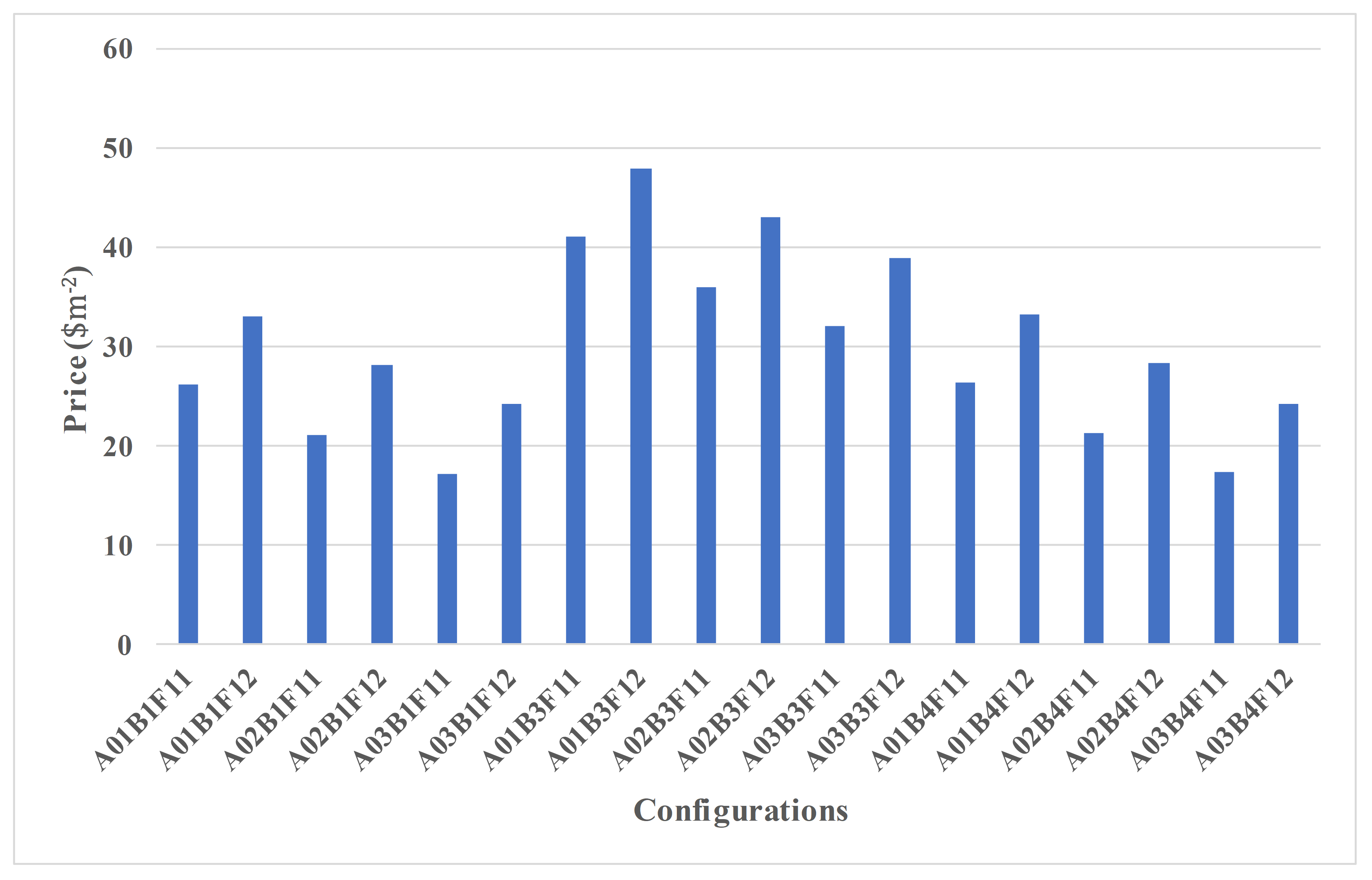}
  \caption{System trade-off analysis among different possible configurations.}
\end{figure}%
According to Figure 2, there are multiple potential design alternatives for developing the robot system. Nevertheless, the constraints of cost and the need for medical-grade testing accuracy suggest specific configurations to be prioritized. The costs associated with ensuring high accuracy are not factored into the trade-off analysis, as they represent stringent limitations. Among the remaining components of the subsystem, a trade-off analysis is conducted, as depicted in Figure 3. This illustrates the expenses associated with each potential design option. The material costs are calculated in USD, with the pricing based on the cost per square meter of the construction materials.  
\subsection{Configuration Selection}
Among different configurations of Figure 3, configurations 'A03B1F11' and 'A03B4F11' entail the lowest cost ($\$m^{-2}$). Consequently, the combination of these two stands as the optimized choice for the system. A combination of these two is used as the final configuration for system development. The final system contains the configuration 'A03A12B1B4C2C4D3E1E4F11' as shown in Figure 2. With these system components, the prototype is developed. 
\subsection{Evaluation and Feedback}
Among the configurations considered, two stand out as optimal choices, A03B1F11 and A03B4F11, as they strike the right balance in system trade-offs. The combined use of these configurations effectively meets the specified constraints. Notably, in Figure 2, the evaluation of availability and specification F12 proves to be better suited for the robot. While F11 is more economical, F12 exhibits superior performance under load conditions. Additionally, in terms of capacity increment complexity, F12 emerges as a practical and viable option. Considering the marginal increase in cost, this approach ultimately yields the optimal solution, culminating in the selection of the final configuration 'A03B1B4F12'.

\section{Prototyping}
\begin{figure}
  \centering\includegraphics[width=\linewidth,trim={0mm 0mm 90mm 0mm},clip]{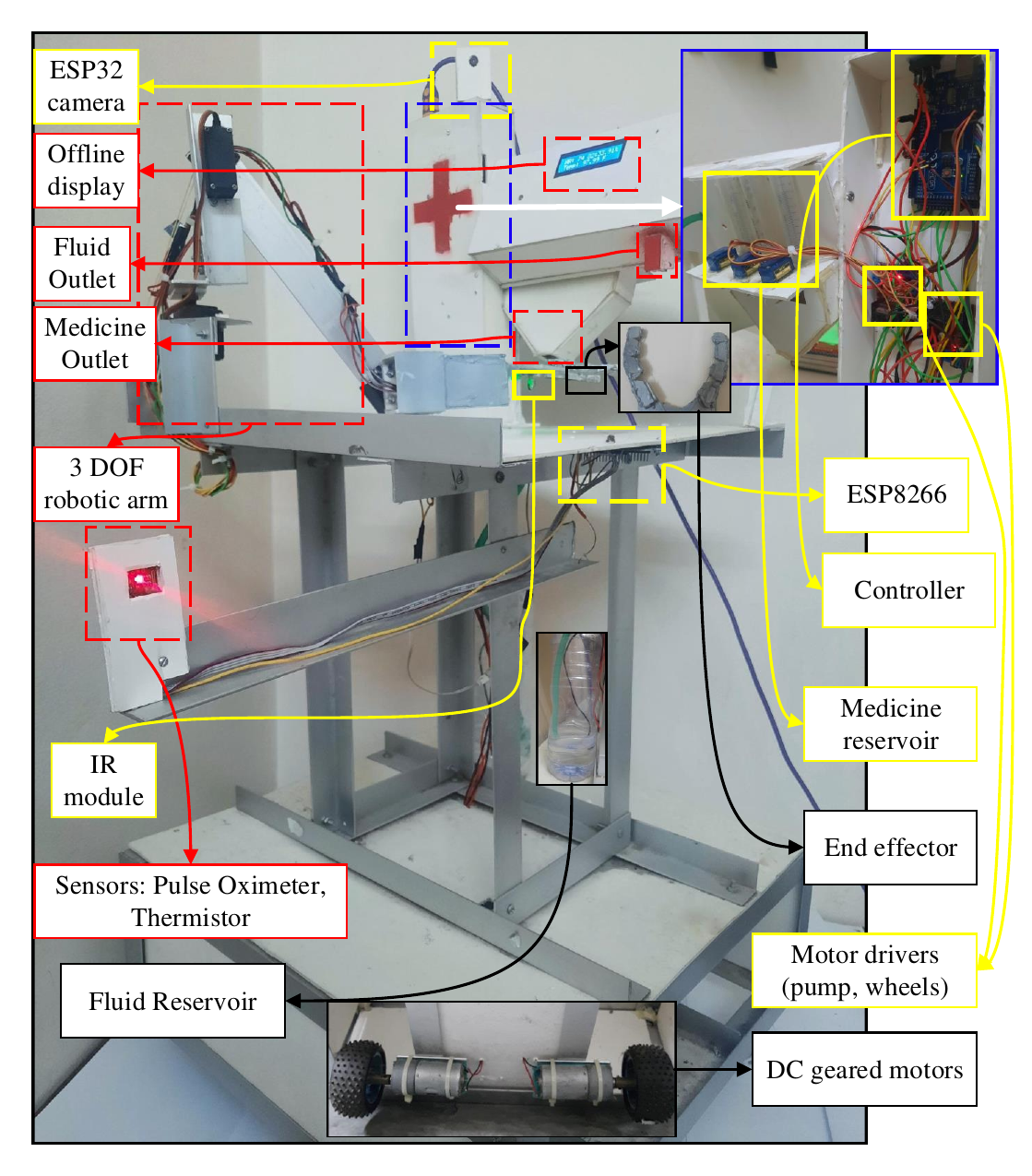}
  \caption{Overall setup for data acquisition, operation, and control tests.}
\end{figure}%

\subsection{Motion Planning}
The system prototype is presented in Figure 4. Here, a DC-geared motor is used for the movement of the robot. The motors are connected to a motor driver which is used to vary the PWM signal. For handling speed synchronization in real-time sensor-based encoder is used which measures both the speed ($\Pi_{1,2}$) and adjusts them to the desired speed $\Pi_{d,c}$. The speed is specified within the maximum limit. A PID controller is designed to minimize the error parameters $\epsilon_{1,2}$.

\begin{equation}
    \epsilon_1=\Pi_{d,c}-\Pi_1
\end{equation}
\begin{equation}
    \epsilon_2=\Pi_{d,c}-\Pi_2
\end{equation}

The controller produces PWM signal $\phi_{1,2}$ from the following equations.

\begin{equation}
\phi_1[\jmath] = K_{1p} \cdot e_1[\jmath] + K_{1i} \cdot \sum_{i=0}^{\jmath} e_1[i] \cdot T_s + K_{1d} \cdot \frac{e_1[\jmath] - e_1[\jmath-1]}{T_s}
\end{equation}

\begin{equation}
\phi_2[\jmath] = K_{2p} \cdot e_2[\jmath] + K_{2i} \cdot \sum_{i=0}^{\jmath} e_2[i] \cdot T_s + K_{2d} \cdot \frac{e_2[\jmath] - e_2[\jmath-1]}{T_s}
\end{equation}
For the discrete time PID controller $T_s$ represents sampling time and $\jmath$ represents time step. The manual encoder design supports cost constraints in terms of encoder-integrated motors. Another DC motor embedded with the fluid pump is used with a separate motor driver and fixed PWM signal.

\subsection{Data Acquisition}
The sensing process in a pulse oximeter using the MAX30101 sensor involves emitting both red and infrared light toward a sample, typically a finger. As the light penetrates the tissue, it interacts with blood vessels. Hemoglobin in the blood absorbs a portion of the light, the amount of which depends on its oxygen saturation and concentration. The sensor then detects the light that is transmitted through the tissue and reflected back. This includes both the DC (constant) and AC (pulsatile) components for both red and infrared light. These raw data values ($\digamma$ values) are processed to isolate the pulsatile AC components. The AC component of the red light signal is analyzed to estimate the heart rate, often by detecting peaks and measuring the time between them. Using the AC components of both red and infrared light, along with the Beer-Lambert law (10), oxygen saturation (SpO2) is calculated, providing a non-invasive estimate of blood oxygen levels. According to the Beer-Lambert law, the absorbance ($A_b$) is directly proportional to the concentration ($c_t$) of the absorbing species and the path depth ($l_d$).

\begin{equation}
A_b = \varepsilon_{ma} \cdot c_t \cdot l_d
\end{equation}
The $\varepsilon_{ma}$ represents a constant name molar absorptivity. The molar Absorptivity of Oxygenated Hemoglobin ($HbO_2$) at around red light (660 nm) is approximately $1.5 \times 10^4 L/(mol \cdot cm)$ and is approximately 
$8.83 \times 10^3 L/(mol \cdot cm)$ for IR (940 nm). With these values, the oxygen saturation ($SpO_2(\%)$) of blood is calculated as (11).

\begin{equation}
SpO_2 = 100\% \times \left( \frac{{\varepsilon_{\text{red}} \cdot l \cdot \digamma_{\text{AC\_red}}}}{{\varepsilon_{\text{IR}} \cdot l \cdot \digamma_{\text{AC\_IR}} + \varepsilon_{\text{red}} \cdot l \cdot \digamma_{\text{AC\_red}}}} \right)
\end{equation}

Here,
\begin{align*}
\digamma_{\text{AC\_red}} &: \text{AC component of red light intensity.} \\
\digamma_{\text{DC\_red}} &: \text{DC component of red light intensity.} \\
\digamma_{\text{AC\_IR}} &: \text{AC component of infrared light intensity.} \\
\digamma_{\text{DC\_IR}} &: \text{DC component of infrared light intensity.} \\
\end{align*}

Likewise, the pulse corresponds to the AC peaks detected by the sensor. These peaks indicate the electrical variation in the physical signal's response. The period of the AC signal ($T_{2\pi}$) is computed and in accordance with (12), the heart rate ($H_R$) is obtained.

\begin{equation}
H_R (BPM) = \frac{60}{T_{2\pi}}
\end{equation}

For sensing body temperature, a thermistor is used. The thermistor's resistance (\(R_2\)) is determined through a voltage divider equation, where \(R_s\) is the known reference resistance (\(10,000\) ohms) and \(V_o\) represents the analog voltage reading from the thermistor.
\begin{equation}
    R_2 = R_s \times \left(\frac{1023.0}{V_o} - 1.0\right)
\end{equation}

The logarithm of \(R_2\) (\(\log(R_2)\)) is subsequently calculated. The Steinhart-Hart equation is employed to compute the temperature (\(T\)) in Kelvin, using coefficients \(c_1\), \(c_2\), and \(c_3\) along with the logarithm of \(R_2\) (\(\log(R_2)\)). Here, \(c_1\), \(c_2\), and \(c_3\) are constants that characterize the behavior of the thermistor.
\begin{equation}
    T_k = \frac{1}{{c_1 + c_2 \cdot \log(R_2) + c_3 \cdot \left(\log(R_2)\right)^3}}
\end{equation}

Finally, the temperature in Kelvin is converted to Fahrenheit using a simple formula. The resulting temperature in Fahrenheit (\(T_F\)) is then displayed on the serial monitor. This code employs these mathematical relationships to accurately convert the analog input from the thermistor into a temperature reading.
\begin{equation}
   T_F = \frac{T_k \times 9.0}{5.0} + 32.0 
\end{equation}

\subsection{Robotic Arm}
For supplying the medicines and necessary fluid a 3 DOF planer manipulator is designed. The manipulator holds the medication pots with a flexible gripper. The arm design parameters and designed arm are shown in Figure 5.

\begin{figure}
  \centering\includegraphics[width=\linewidth,trim={30mm 20mm 70mm 50mm},clip]{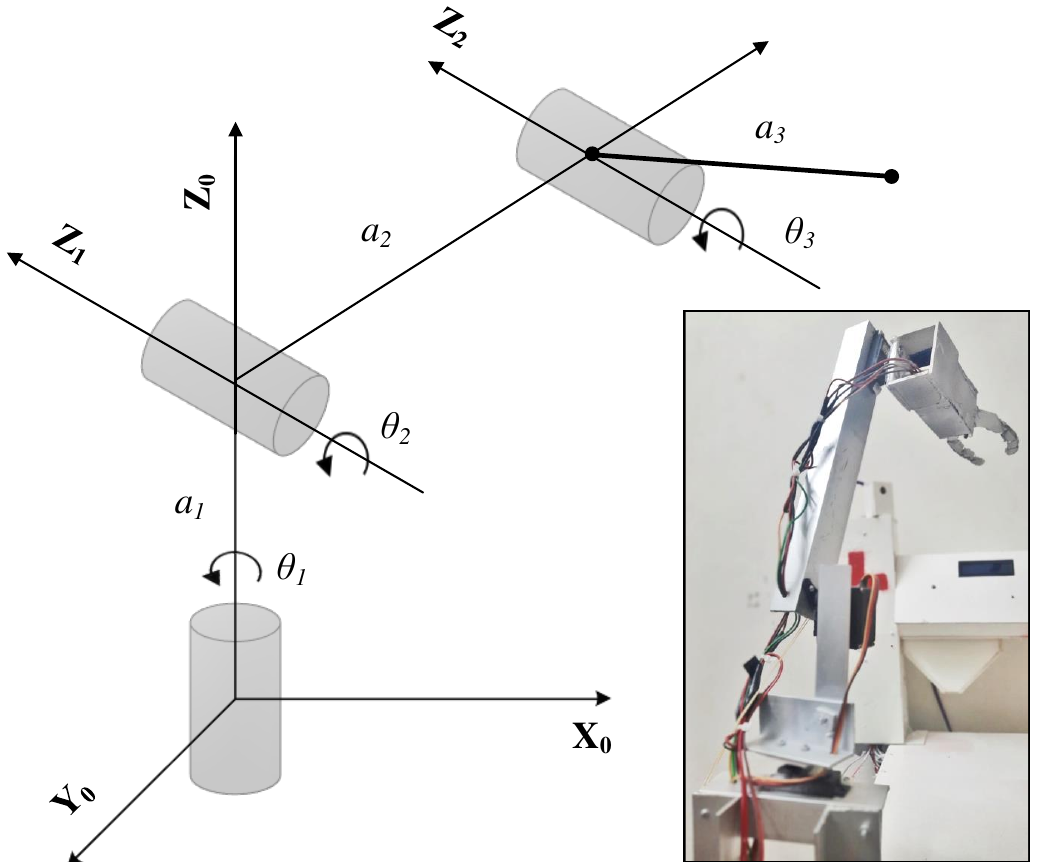}
  \caption{Schematic of the 3 DOF planer mechanism.}
\end{figure}%

Table 2 lists the DH parameters for the 3-link manipulator, including joint angles (\(\theta_i\)), link offsets (\(d_i\)), link lengths (\(a_i\)), and link twists (\(\alpha_i\)). 

\begin{table}[ht]
    \centering
    \caption{Denavit-Hartenberg (DH) Parameters}
    \begin{tabular}{|c|c|c|c|c|}
        \hline
        \(i\) & \(\theta_i\) & \(d_i\) & \(a_i\) & \(\alpha_i\) \\
        \hline
        1 & \(\theta_1\) & 0 & 10 cm & 0 \\
        2 & \(\theta_2\) & 0 & 22 cm & 0 \\
        3 & \(\theta_3\) & 0 & 15 cm & 0 \\
        \hline
    \end{tabular}
\end{table}

Following this, the transformation matrix \(A_i\) is defined using trigonometric functions and these DH parameters, which relate the coordinates of one link to the previous link.
\begin{equation}
    A_i = \begin{bmatrix}
    c(\theta_i) & -s(\theta_i)c(\alpha_i) & s(\theta_i)s(\alpha_i) & a_ic(\theta_i) \\
    s(\theta_i) & c(\theta_i)c(\alpha_i) & -c(\theta_i)s(\alpha_i) & a_is(\theta_i) \\
    0 & s(\alpha_i) & c(\alpha_i) & d_i \\
    0 & 0 & 0 & 1
\end{bmatrix}
\end{equation}

The overall transformation matrix \(T\) is then calculated by multiplying the individual \(A_i\) matrices. Additionally, the Jacobian matrix \(J\) is defined, which relates the end-effector's coordinates and orientations to the joint velocities.
\begin{equation}
    T = A_1 \cdot A_2 \cdot A_3
\end{equation}
In creating this robot arm, we rely on the Jacobian matrix, denoted as \(J\). The coordinates \(x, y, z\) stand for the position of the end effector in three-dimensional space, while \(\theta_1, \theta_2, \theta_3\) correspond to the joint angles of the arm. The terms \(\frac{\partial x}{\partial \theta_1}, \frac{\partial y}{\partial \theta_1}, \frac{\partial z}{\partial \theta_1}\) signify how a minute change in the first joint angle \(\theta_1\) influences the respective alterations in the x, y, and z coordinates of the end effector. Similarly, \(\frac{\partial x}{\partial \theta_2}, \frac{\partial y}{\partial \theta_2}, \frac{\partial z}{\partial \theta_2}\) and \(\frac{\partial x}{\partial \theta_3}, \frac{\partial y}{\partial \theta_3}, \frac{\partial z}{\partial \theta_3}\) represent the effects of changes in the second and third joint angles, respectively. To transfer the medicines from a specific medicine outlet to the patient desk coordinate helps the end effector state estimation via joint angles.

\begin{equation}
    J = \begin{bmatrix}
    \frac{\partial x}{\partial \theta_1} & \frac{\partial x}{\partial \theta_2} & \frac{\partial x}{\partial \theta_3} \\
    \frac{\partial y}{\partial \theta_1} & \frac{\partial y}{\partial \theta_2} & \frac{\partial y}{\partial \theta_3} \\
    \frac{\partial z}{\partial \theta_1} & \frac{\partial z}{\partial \theta_2} & \frac{\partial z}{\partial \theta_3} \\
\end{bmatrix}
\end{equation}

 For a new coordinate input ($P_d(x,y,z)$), the position is gained from the inverse kinematics approach. Firstly, \(\theta_1\) is computed as the arc-tangent of the ratio of \(y_d\) to \(x_d\), determining the angle of the first joint. 
 \begin{equation}
     \theta_1 = \arctan2(y_d, x_d) 
 \end{equation}
 
 Next, \(d_3\) is calculated as the difference between the desired \(z\)-coordinate (\(z_d\)) and the fixed offset \(d_1\), establishing the length of the third link.
 \begin{equation}
     d_3 = z_d - d_1
 \end{equation}
 
 The angle \(\theta_2\) is determined by finding the arc-tangent of the square root of the sum of the squares of \(x_d\) and \(y_d\) minus \(a_1\) with respect to \(d_3\), and then subtracting the arc-tangent of \(a_2\) over \(d_3\). This yields the angle of the second joint.
 \begin{equation}
    \theta_2 = \arctan2(\sqrt{x_d^2 + y_d^2} - a_1, d_3) - \arctan2(a_2, d_3) 
 \end{equation}
 
 The Jacobian \(J\) is composed of partial derivatives that relate the end-effector coordinates to the joint angles. The pseudo-inverse \(J^-\) is calculated to map desired end-effector velocities to the necessary joint velocities. Finally, the joint velocity vector \(\dot{\theta_1}, \dot{\theta_2}, \dot{\theta_3}\) is determined by multiplying \(J^-\) by the desired end-effector velocity vector \(\dot{x}, \dot{y}, \dot{z}\), enabling dynamic control of the robot's motion.
\begin{equation}
    J^- = (J^T J + \lambda^2 I)^{-1} J^T
\end{equation}
\begin{equation}
    \begin{bmatrix}
    \dot{\theta_1} \\
    \dot{\theta_2} \\
    \dot{\theta_3}
\end{bmatrix} = J^- \begin{bmatrix}
    \dot{x} \\
    \dot{y} \\
    \dot{z}
\end{bmatrix}
\end{equation}
\subsection{Medication System}
The RoboNurse system is equipped with a sophisticated medication delivery system, featuring three distinct cylinders each containing a different type of medicine. To regulate access to these medicines, three servo motors have been employed to efficiently open and close the respective cylinder gate valves. The control signals for these servos are orchestrated by a controller, which operates based on a knowledge base that has been meticulously trained to comprehend various health states. Additionally, manual commands can also be transmitted through an Internet of Things (IoT) interface, providing an extra layer of flexibility in the medication delivery process. This comprehensive system ensures precise and timely dispensing of medicines, tailored to the specific health requirements of the individual receiving the treatment. The schematic of the subsystem is shown in Figure 6.
\begin{figure}
  \centering\includegraphics[width=\linewidth,trim={0mm 0mm 130mm 0mm},clip]{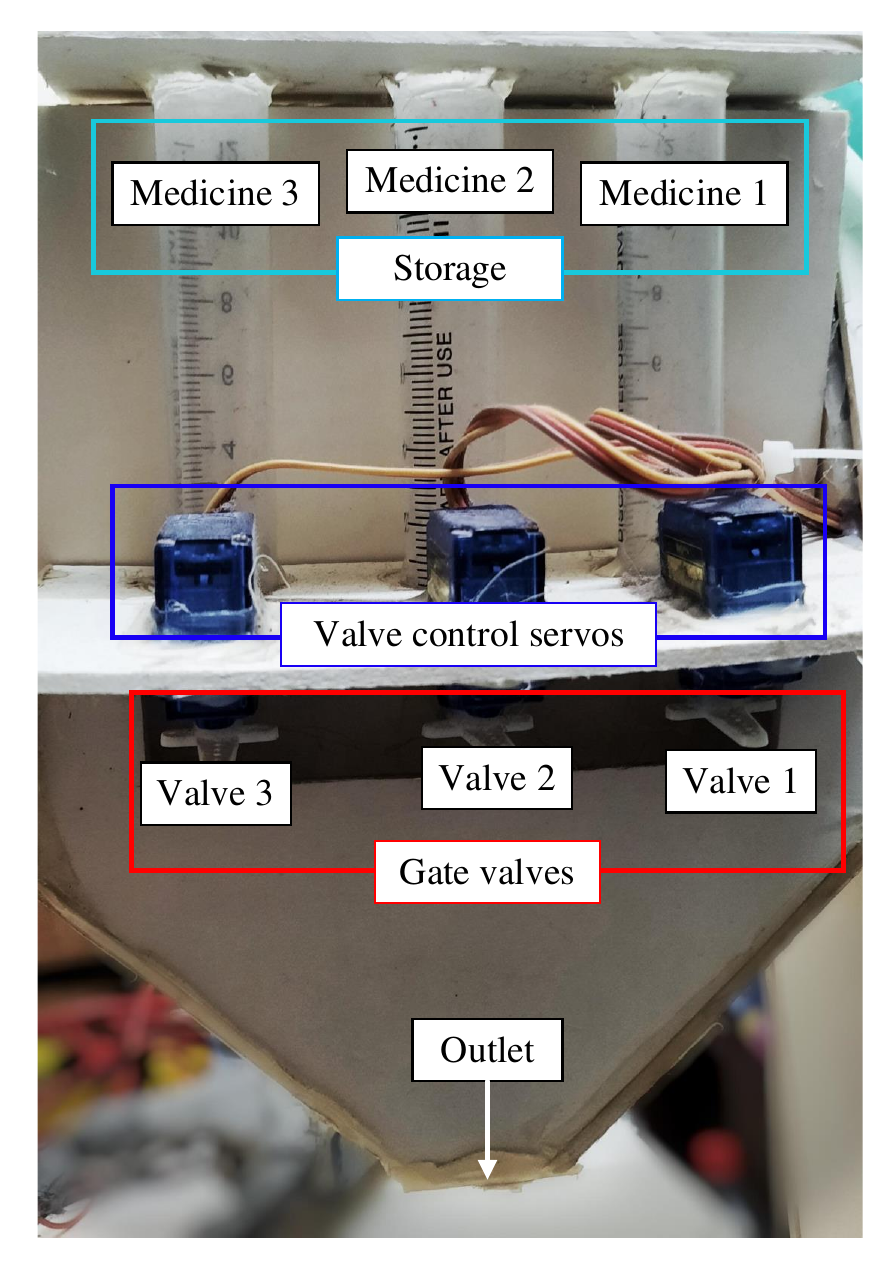}
  \caption{Medicine storage and valve control system.}
\end{figure}%

\subsection{Operation and Control Strategies}
The robot remains stationed at a designated docking base within a specific room $Bxx$. Its operations can be categorized into two distinct types, both of which are outlined in Algorithm 1. The first type involves routine procedures, while the second pertains to supervisory control. For regular check-ups and medication administration, the routine operation scheme is employed. In this mode, the robot adheres to a predetermined schedule. Upon activation, it systematically attends to each patient by following the trajectory function $f_{ij}(t_i)$ stored in its storage. The robot gathers health data ($\xi(\lambda_i)$) from each patient and transmits it to the cloud interface, where the information is archived. With this data, experts can assess the patient's health conditions. Additionally, the system monitors the progress, improvements, and impact of specific medications ($\psi(\gamma_i)$). Based on the assessed health states, the robot dispenses medications ($\gamma_i$) in accordance with the expert system's recommendations. The medication triggering function $\psi(\gamma_i)$ interacts with the control system trained actuation values stored in $\sigma(\rho_i)$, Again, the corresponding medication code is also transmitted to the cloud storage for the respective patient. The robot iterates through this process until it tends to the final patient, after which it returns ($f^{ij}(t_i^{-1})$) to its base for sanitation and replenishment of fluids and supplies.      

\begin{algorithm}
\caption{RoboNurse Control: Routine and supervisory}\label{alg:combined}
\begin{algorithmic}[1] 
\STATE 
\STATE {\textsc{\textbf{TRAIN}}}$(trajectory \leftarrow t_i; health-states, \xi(\lambda_{i});$
\STATE \hspace{1.3cm}$medication/actuation \leftarrow \psi(\gamma_i))$
\STATE {\textsc{Trajectory}},$f^{tj}(t_{i})$
\STATE \hspace{0.5cm}$store\leftarrow f^{tj}(t)=\sum_{i=1}^{n}t_i$
\STATE \hspace{0.5cm}$store\leftarrow t_{i} \hspace{0.5cm} \forall \hspace{0.5cm} Bxx$
\STATE \hspace{0.5cm}$write\rightarrow f^{tj}(t_{i}) \hspace{0.5cm} \forall \hspace{0.5cm} Bxx(routine)$
\STATE \hspace{0.5cm}$write\rightarrow f^{tj}(t_i) \hspace{0.5cm} \in \hspace{0.5cm} B0i(command)$
\STATE {\textsc{\textbf{Read:}}} $\xi(\lambda_{i})$
\STATE \hspace{0.5cm}$\Rightarrow \psi(\gamma_{i}) = \left\{ \begin{array}{rcl}
\gamma_1  & \mbox & \xi\in(\lambda_x,\lambda_y,\lambda_z)\\ \gamma_1 & \mbox & \xi\in(\lambda_x,\lambda_y,\lambda_z) \\ \gamma_3  \end{array}\right\}$
\STATE {\textsc{\textbf{Actuation:}}} $\sigma(\rho_{i})$
\STATE \hspace{0.5cm}$output\rightarrow \psi(\gamma_{i}) \hspace{0.5cm} \forall \hspace{0.5cm} \xi(\lambda_{i})$
\STATE \hspace{0.5cm}$\Rightarrow \sigma(\rho_i)\leftarrow \psi(\gamma_i)$
\STATE \hspace{0.5cm}$write:\hspace{0.3cm} \sigma(\rho_i) \hspace{0.5cm} \forall \hspace{0.5cm} \psi(\gamma_i)$
\STATE
\STATE \textbf{Supervisory Control}
\STATE {\textsc{Read}} $(DockTimer(T) ;command(S^t));$
\STATE \hspace{0.5cm}while$S^t_0\longleftarrow S^{t++}_i \hspace{0.3cm} \forall \hspace{0.3cm} B0i$
\STATE \hspace{0.5cm}select:$f^{tj}(t_i)\rightarrow$ write sequence $\hspace{0.2cm}\in\hspace{0.2cm}B0i$
\STATE {\textsc{Get:}} $\xi(\lambda_{i}) \hspace{0.2cm} while \hspace{0.2cm} S^t_0\leftarrow S^{t++}_i$
\STATE \hspace{0.5cm}$display\hspace{0.2cm}\longrightarrow \lambda_i \hspace{0.2cm} \in \hspace{0.2cm} B0i$
\STATE \hspace{0.5cm}$reset:\hspace{0.2cm} S^t_i\longrightarrow S^t_0$
\STATE \hspace{0.5cm}$read:\hspace{0.2cm} S^t_0\longleftarrow S^t_i$
\STATE \hspace{0.5cm}$return:\hspace{0.2cm} f^{tj}(t_i)\longrightarrow f^{tj}(t_i^{-1})$
\STATE \hspace{0.5cm}$write:\rightarrow \sigma(\rho_{i}) \hspace{0.5cm} \in \hspace{0.5cm} S_i^t(command)$
\STATE \hspace{0.5cm}$return:\hspace{0.2cm} f^{tj}(t_i)\longrightarrow f^{tj}(t_i^{-1})$
\STATE 
\STATE \textbf{LOG:}
\STATE \hspace{0.5cm}$health-states \leftarrow \xi(\lambda_{i})$
\STATE \hspace{0.5cm}$medication \leftarrow \xi_i^t$
\end{algorithmic}
\end{algorithm}

\section{Results}
The system is structured with a focus on functionalities such as conducting health assessments and initiating medication protocols based on the individual's health status. It incorporates both standard automated functions and allows certain operations to be conducted remotely through a cloud-based interface. Additionally, the system is equipped with essential safety measures, including obstacle handling, the ability to adjust schedules through commands, and the activation of emergency response procedures. These elements are implemented with a keen eye on cost-effectiveness. Consequently, the results section not only encompasses the achieved functionalities but also evaluates the robot's overall performance.

\subsection{Health Care}
\begin{figure}
  \centering\includegraphics[width=\linewidth,trim={30mm 80mm 30mm 90mm},clip]{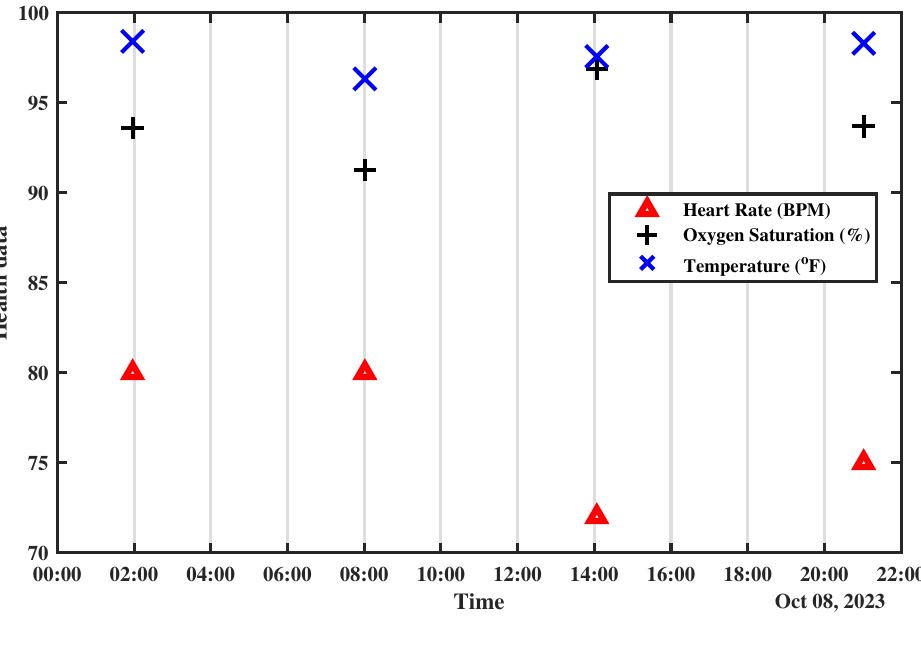}
  \caption{Routine and command-based checkup data for a patient.}
\end{figure}%

The standard procedure involves gathering health information at a designated time of day. Following an analysis using the knowledge base, the robot proceeds with making decisions regarding medication. Additionally, supervisory tasks occur when a specific patient requires special attention or is in an emergency situation. In instances where priority is assigned through a command-based approach, it can override the routine operation. In cases where manual intervention is necessary, medication or fluid supply can be administered based on the acquired data. This data is depicted in Figure 7. The patient's health metrics, including pulse rate, oxygen saturation, and temperature, are both stored in a cloud-based database and displayed in real-time. Depending on the patient's health status, the medication system is either activated or kept inactive. When a specific medicine (Mxx) is triggered, it prompts the opening of its respective reservoir valve, allowing the medicine to be dispensed. The sequence of these triggers is illustrated in Figure 8.     

\begin{figure}
  \centering\includegraphics[width=\linewidth,trim={30mm 80mm 30mm 130mm},clip]{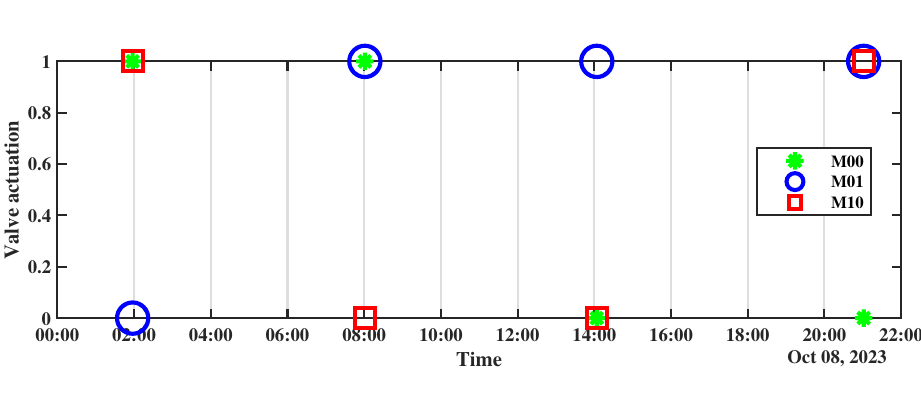}
  \caption{Medication system valve triggering sequence for the routine operation.}
\end{figure}%

The setup engages in bidirectional data and command exchange with the cloud layer, ensuring seamless, live-time communication. The Arduino IoT cloud interface is employed to present the acquired data visually, striking a balance between cost-effectiveness and functionality. The real-time data interface is shown in Figure 9. The cloud layer not only displays this real-time information but also archives it, facilitating the monitoring of patients' progress and enabling the analysis of data for future reference. This, in turn, aids in comprehending the effects of various medications on patients and contributes to the accumulation of a comprehensive database for monitoring pandemic patterns.

\begin{figure}
    \centering
    \includegraphics[width=\linewidth]{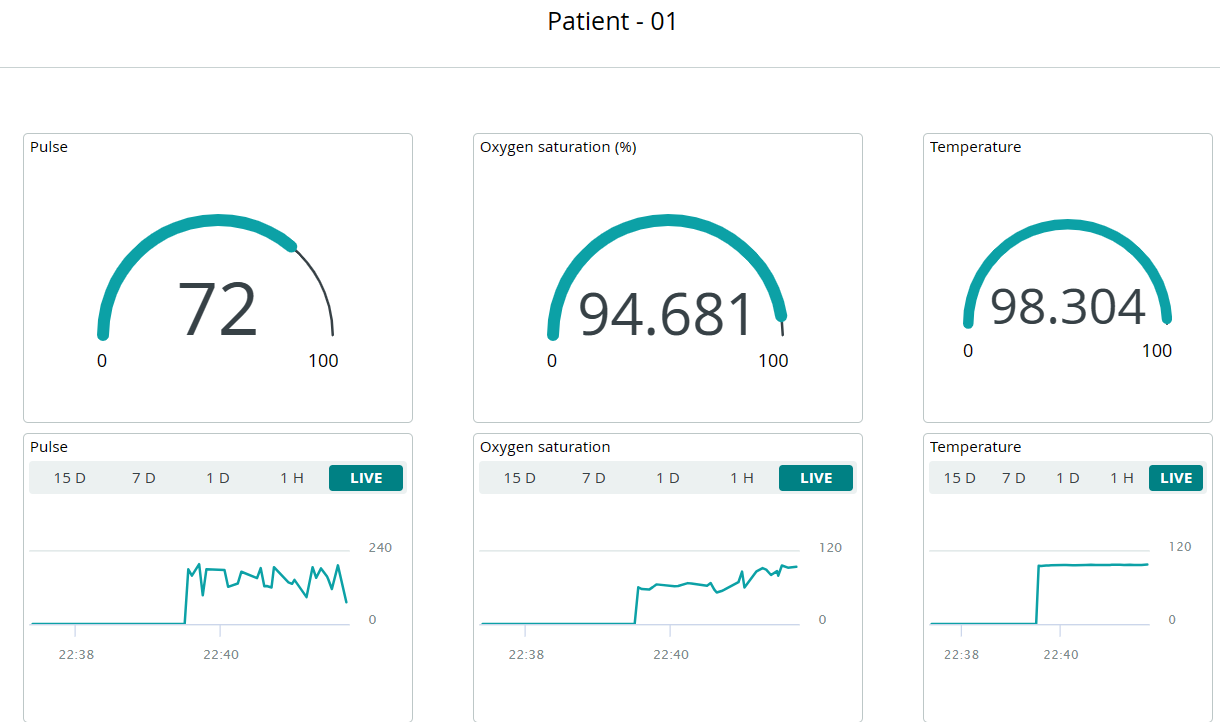}
    \caption{Cloud data acquisition and visualization interface.}
    \label{fig:enter-label}
\end{figure}

\subsection{Vision System}
The robot employs a camera module chosen during the configuration selection process. This module is chosen to balance cost with essential functionalities needed for the robot nurse system. The vision system serves to monitor the patient's physical condition and enables remote communication. The viewing angle of the camera in use is illustrated in Figure 10. Prior to implementation, the camera underwent calibration to ensure it captured the widest possible viewing angle. Additionally, the vision system allows for observing arm movements during manual control. The camera's angle can be adjusted by about 30 degrees in both directions through the servo mechanism on which it is mounted.

\begin{figure}
  \centering\includegraphics[width=\linewidth,trim={20mm 75mm 110mm 20mm},clip]{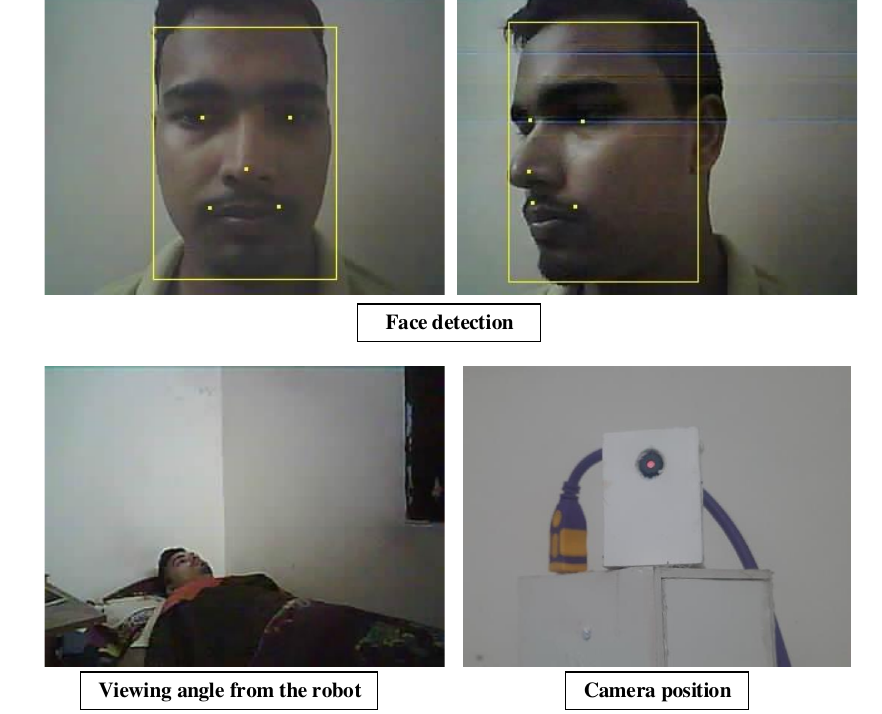}
  \caption{Robot nurse's visual system for monitoring patients and overseeing the movement of the supervisory actions.}
\end{figure}%
\subsection{Performance Parameters}
The system employs a systems engineering approach during its development, involving trade-offs among potential configurations. Given the system's objective of creating a cost-effective healthcare solution with the highest attainable precision, it becomes imperative to monitor its performance closely. This emphasis on function over form in robot design, as seen in Table 4, yields the observed results. The robot's maximum speed is determined by the motor's capability, even after multiple usage cycles, and any degradation in rotational output is regulated using a feedback-based PID controller. An IR module serves as the feedback signal source, aiding in error calculation for speed adjustment. Additionally, the robot maintains continuous communication with a cloud database, leading to variable cloud layer communication latency, ranging from 500 ms to 1200 ms, contingent on network quality. However, the cloud data update interval remains consistent based on the code. An 1100 ms value is set for sensors to collect sufficient samples prior to uploading, while response time varies according to network quality. The performance data are collected with multiple iterations to deal with the lifecycle decay hence ensuring reliability.   

\begin{table*}
    \caption{Performance index of the developed robot nurse.}
    \centering
    \begin{tabular}{p{5cm}p{2cm}|p{5cm}p{2cm}}\hline
        Entity & Data & Entity & Data \\
        \hline
        Avg. motor speed & 390 RPM & Serial communication delay & 36 ms \\
        Robot speed & 1.74 m/s & Cloud update delay & 900 ms \\
        Arm motor speed & 5.24 rad/s & Cloud update period & 1100 ms \\
        Arm wt. capacity & 280 gm & Max. No. of medicines & 3 \\
        Robot wt. capacity & 15 Kg & Avg. medication time & 2.88 s\\
        Fluid pump capacity & 96.42 lt/hr & Avg. checkup time  & 28.07 s\\
        Battery life (running) & 1.38 hrs & Avg. response time  & 1.16 s\\
         \hline
    \end{tabular}
\end{table*}
\section{Conclusion }
This article has successfully designed and tested a robot nurse with a primary focus on patient health monitoring and precise medication administration, all while maintaining a low development cost. The robot records the health status and medication details for analyzing the response to specific medicines. Its operations can be categorized into two systems: supervisory and autonomous control. In the autonomous mode, the robot administers medications based on its trained knowledge base. In the supervisory mode, the robot is controlled through teleoperation using remote commands.

The development process employs a systems engineering approach to optimize costs. This involves breaking down the system into various subsystem components and comparing costs for different configurations. Through trade-off analysis, a low-cost pathway is chosen and implemented during development. The system operates within two constraints: cost is a predefined limit, and accuracy is a constraint within the domain. Component selection takes both of these constraints into consideration. The final system is not only cost-effective but also maximizes the accuracy of medication administration through training. The system performance is assessed which ensures the reliability of repetitive performance.

However, there are areas that could benefit from further development in future work, which include:
\begin{enumerate}
    
\item Enhancing the image processing system to improve accuracy, ensuring the precise placement of oxygen masks.

\item Adapting the expert system to accommodate situations, where a higher volume of routine medications may be required.

\item Creating a sensory retrofit for human wearables that can transmit health status data to both a cloud-based database and the robot, enabling timely responses in case of emergencies.
\end{enumerate}
\medskip

%
\bibliography{cas-refs}   
\vfill

\end{document}